\documentclass[conference]{IEEEtran}
\IEEEoverridecommandlockouts

\usepackage{natbib}
\usepackage{amsmath,amssymb,amsfonts}
\usepackage{graphicx}
\usepackage{textcomp}
\usepackage{xcolor}
\usepackage[hidelinks]{hyperref}
\usepackage{algorithm}
\usepackage{algpseudocode}

\hypersetup{
    colorlinks=true,
    linkcolor=blue,
    filecolor=blue,      
    urlcolor=blue,
    citecolor=blue
}

\def\BibTeX{{\rm B\kern-.05em{\sc i\kern-.025em b}\kern-.08em
    T\kern-.1667em\lower.7ex\hbox{E}\kern-.125emX}}
\begin{document}

\title{Generating Bayesian Network Models from Data Using Tsetlin Machines}

\author{
\IEEEauthorblockN{Christian Blakely}\thanks{Author's status: {Any statements, ideas, and/or opinions expressed in this paper are strictly those of the author and do not necessarily reflect those of PwC Switzerland}}
\IEEEauthorblockA{\textit{Machine Learning and AI} \\
\textit{PwC Switzerland}\\
Zurich, Switzerland\\
christian.blakely@pwc.ch}}

\maketitle

\begin{abstract}
Bayesian networks (BNs) are directed acyclic graphical (DAG) models that have been adopted into many fields for their strengths in transparency, interpretability,
probabilistic reasoning, and causal modeling. Given a set of data, one hurdle towards using BNs is in building the network graph from the data that properly handles 
dependencies, whether correlated or causal. In this paper, we propose an initial methodology for discovering network structures using Tsetlin Machines (TMs). 
\end{abstract}


\section{Introduction}

A Bayesian network (BN) is a directed acyclic graph (DAG) where nodes of the graph are random
variables while the directed edges represent dependent relationships in the form of probability dependencies among 
the variables. The entire graph thus represents a compact representation of a joint probability distribution over the set of random variables, where each node and its parents are associated with a conditional probability distribution (CPD). Reasons for adopting Bayesian network models over other machine learning models such as random forests, gradient-boost models, or neural networks usually aim towards 1) Interpretability 2) Uncertainty Quantification 3) Small Data Requirements, and 4) efficient computational complexity. Furthermore, since the entire joint probability distribution is described completely by the network, generating new data from the network is also a common application.

One challenge of modeling with BNs is in constructing this DAG that accurately and robustly characterizes the underlying 
relationships in data \cite{scutari2010learning}. Learning a BN is typically conducted in two phases, whereby the first phase constructs the topology (network structure) of the network and the second phase estimates the parameters of the CPDs given the fixed structure. As parameter estimation is considered a well-studied problem, especially with the availability of data, learning the structure is more challenging. A long literature study of BN structure learning includes usually two different types of algorithms: 1) constraint-based which look for conditional independencies (CIs)in the data and
build the DAG consistent to these CIs and 2) score-based which convert the problem to an optimization problem, where a score is used to measure the fitness of a DAG to the observed data with a search strategy employed to maximize the score over the space of possible DAGs (see e.g. \cite{scutari2015bayesian} for more on constraint-based algorithms, and \cite{kasza2011comparison} for more on score-based approaches). No matter the methodology or combination of methodologies employed, finding an optimal structure has been shown to be NP-hard. In light of this, we will focus on the problem of generating network structures for the data, as defining CPDs with the use of data is a much easier problem once conditional dependencies have been defined.  

\subsection{Motivation and contributions}

In the past few years, there has been a new genre of building supervised machine learning models called Tsetlin machines. 
First introduced by O.C. Granmo in \cite{granmo2018}, Tsetlin machines have been gaining attraction in some machine learning communities for their many computational and modeling benefits. Mostly notably, TMs capture patterns in data using conjunctive clauses in propositional logic and are thus intrinsically interpretable learning systems, and they are also typically fast to train, benefiting from parallelization and many recent regularization advancements in training algorithms such as in \cite{abeyrathna2021integer}, \cite{dctm}, even without the use of GPUs. 

But the most useful benefit of Tsetlin machines for this study presented in this paper is the ease in which 
global importance features can be extracted from the propositional logic of an underlying learned Tsetlin machine on 
data. In \cite{pwc} a simple method for extracting global importance features was introduced and compared with popular approaches such as SHAP for gradient boost models showing comparable results without the need of a post-hoc approach. 
We take advantage of this powerful approach to derive a fairly straightforward method for constructing Bayesian network topologies of data.

The motivations of learning Bayesian Network structures from data using TMs is multifold:
\begin{itemize}
\item Bayesian Networks can offer a compact visual overview of data dependencies 
\item Through a very powerful reinforcement learning mechanism, TMs can often learn literal structure already with small data
\item Contributes to the validation and verification of TMs in building predictive models
\item Provides a way of building hybrid predictive models
\end{itemize}

We will touch on all these points throughout this paper, with an exposition of the approach using both algorithms and examples. The main contribution of this paper is to demonstrate that TM models can provide an alternative approach to building and investigating Bayesian network structures, or at minimum provide additional computational medium for a hybrid-type approach using other classical resources. 

\subsection{Paper structure}

The paper will be organized as follows. We first review Tsetlin Machines with a focus on constructing interpretable features of the data from the resulting propositional logic expressions that are
learned in training. We then provide a simple algorithm for constructing a Bayesian Network given (unlabeled) data using a modified version of the standard TM learning algorithm. 
To explore validation of the method, we will propose an approach where we first sample from a known Bayesian network and then apply the algorithm to the sampled data to build a network and verify how similar the networks are. We then proceed to a larger statistical approach to understand how well our approach does in generating networks that are close to the ground truth. To understand the proximity to the ground truth network, we apply a kernel graph based approach to compare a TM generated network with the proposed network from which the underlying data was simulated. We do this on varying number of node size networks, and also on varying the models used in generating the networks. 

Finally, in the last sections, we apply our approach to two real-world data sets, namely a Supply Chain data set and a Bank Customer Churn data set from an anonymous bank. The motivation here is to understand how well our method does in building partial dependencies on certain features in the data which are well known in other models for such types of industrial data sets. Furthermore, in the Supply Chain dataset, only 100 samples are available, with over 20 features, making it very challenging for traditional machine learning predictive models such as neural networks and gradient boost approach which typically require lots of data.

%

\section{Tsetlin Machine Architecture}

\begin{figure}[!ht]
\centering
\includegraphics[width=.45\textwidth]{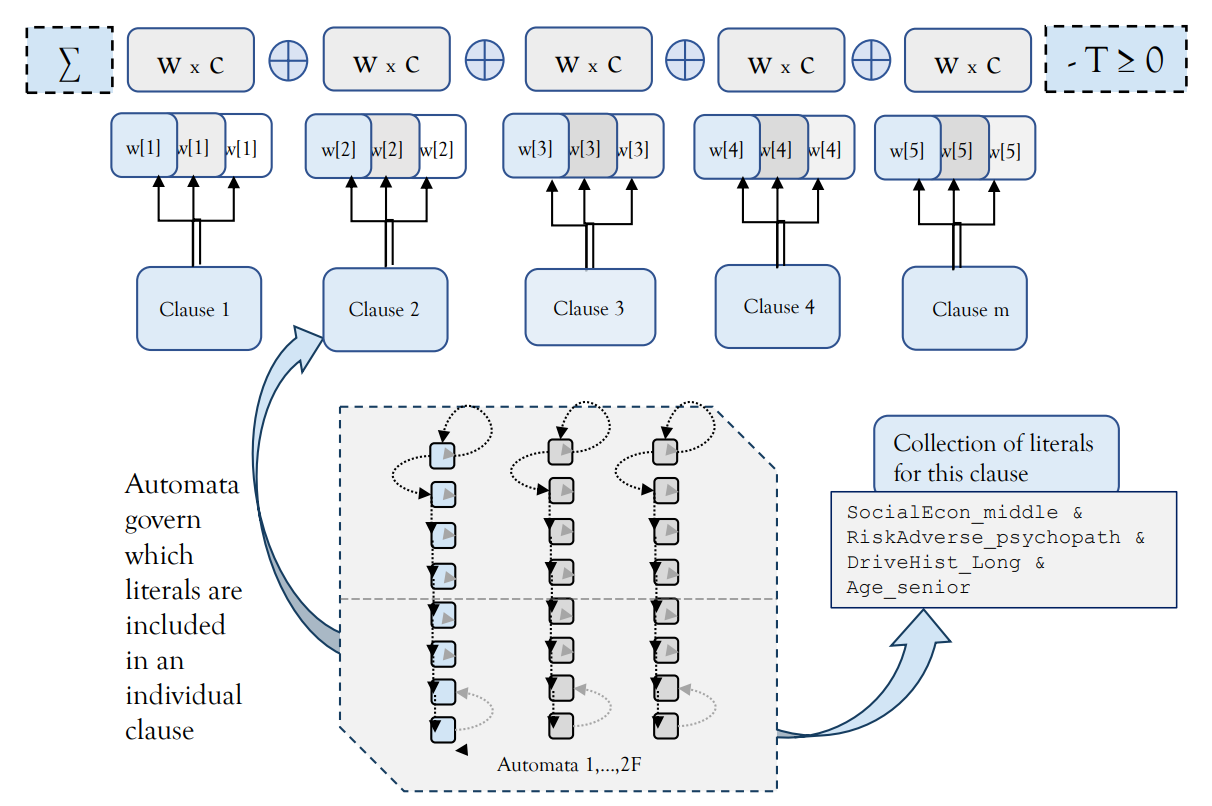}
\caption{Showing a collection of clauses and their relationship with weights that are learned during TM training. Each clause contains a set of literals that are also learned during feedback.}
\label{ref:clause-weights}
\end{figure}

In this paper we employ a recently developed architecture for TM learning originally proposed in \cite{glimsdal2021coalesced}. The architecture shares an entire pool of clauses between all output classes, while introducing a system of weights for each class type. The learning of weights is based on increasing the weight of clauses that receive a Type Ia feedback (due to true positive output) and decreasing the weight of clauses that receive a Type II feedback (due to false positive output). This architectural design allows to determine which clauses are inaccurate and thus must team up to obtain high accuracy as a team (low weight clauses), and which clauses are sufficiently accurate to operate more independently (high weight clauses). The weight updating procedure is given in more detail in Appendix \ref{appendix:weighted}. Here we illustrate the overall scheme in Figure \ref{ref:clause-weights}. Notice that each clause in the shared pool is related to each output by using a weight dependant on the output class and the clause. The weights that are learned during the TM learning steps and are multiplied by the output of the clause for a given input. Thus clause outputs are related to a set of weights for each output class (in this case three classes). One of the clauses also contains four literals, the pattern learned for that clause. 

In our methodology for building candidate networks, both the frequency of literals across the various clauses and the weights will play an important role in determining the global feature strength in learning data outputs, which we give a brief overview of next.   

\section{Global Interpretability}

In this section we give a brief overview of the expressions for global interpretability in TMs that have been modified from \cite{pwc}. In this paper, global strength indicators for each feature were derived based on output class prediction. This required constructing the strength based on the polarity of the clause as we wanted to understand strength of a feature dependent on an output of a class. 

We represent a set of $d_f$ literals for the $i$-th feature $f_i$ as $S_{f_i} = (l_1, l_2, \ldots, l_{d_f})$.   
and map all the literals from all the features of the data into one macro set of literals which we will denote as $F_L = \cup_i S_{f_i}$.
Thus the set of literals $I^n_c$  learned from any TM learning process for class output $n$ and clause $c$ is a subset of $F_L$.  To formulate our global feature importance 
expressions, associated with each clause $c$ is an integer weight $w_n[c]$. These weights will be used to express the most salient features in predicting an output for a feature 
$f_i$. 

$y^i \in \{y^1, y^2, \ldots, y^n\}$, $y^i = \{0,1\}$ and the upper index $i$, which refers to a particular output variable. For simplicity in exposition, we assume that the corresponding literal index sets $I^i_j, \bar{I}^i_j$, for each output variable, $i = 1, \ldots, n$, and clause, $j = 1, \ldots, m$, have been found under some performance criteria of the learning procedure, described in Section \ref{appendix:weighted}.  With these sets fixed, we can now assemble the closed form expressions. 

We denote by $I^n_c$ the set of literals (combined with negated literals) learned from the TM for class output $n$ and clause $c$.  Associated with each clause $c$ is an integer weight $w_n[c]$, for all clauses $c = 1, \ldots, C$.  Specifically, we compute feature strength for a given output variable and $k$th bit of feature $f_u$ as follows:

\begin{equation}\label{global1}
g[f_i] \leftarrow \sum_n \sum_c  |w^n[c] |_{l_k \in S_{f_i} | l_k \in I^n_c}  
\end{equation}

For any feature $f_i$, this expression sums up the absolute value of the weights $w^n[c]$ over all class outputs $1 \leq n \leq N$ and all clauses $c \in C$ such that the literals belong to both the feature's set of literals $S_{f_i}$ and the clause $c \in C$. Once computed for all features $f_i$ of the data, trained on an output variable $y$, we can assess the most important features in any prediction of a collection of input variables by ranking the values of each $g[f_i]$. In other words, for any given feature, the frequency of its set of literals inclusion across all clauses multiplied by the weight of that clause for the output class polarity governed by the governs its global importance score. This score thus reflects how often the feature is part of a pattern that is important to making a certain class prediction. Notice that we are interested in indices pertaining any polarity (and thus weight) of the clauses $C^n_c(X)$ since these are the clauses that contain references to features that are beneficial for predicting $R_n$.

\begin{figure}[!ht]
\centering
\includegraphics[width=.4\textwidth]{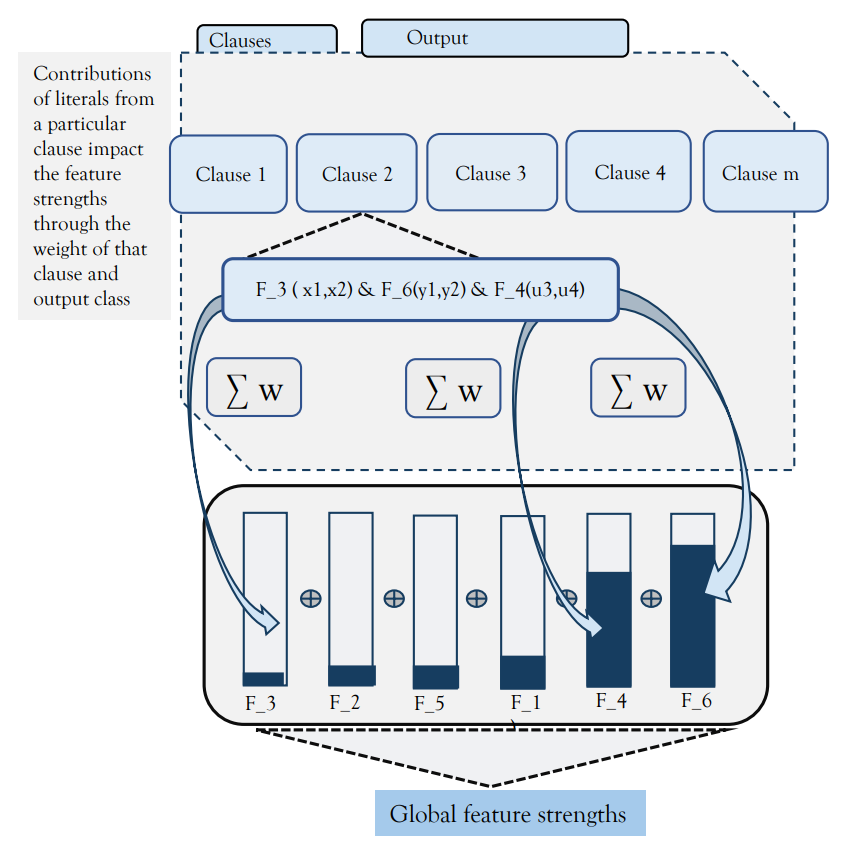}
\caption{Assembling global feature strengths from contributing features in clauses}
\label{ref:outputs}
\end{figure}

Figure \ref{ref:outputs} shows a graphical representation of how the feature strengths are compiled after a TM training round. We look in each clause and count how many literals from the different features exist and place them in a map, incrementing by the absolute value of the weight for that clause. This way, the  highest valued strengths feature the most amount of literals spread out over as many clauses. The weights matter because they are summed across the frequencies of literals. A very small weight compared to the others will contribute very little (as that clause made little contribution to a prediction). More relevant clauses (and thus literals), will be more represented with the larger weights. 

This gives us quick access to feature strength in absolute terms without regard to any output prediction class, which, as we'll see is relevant for when finding possible parent or child nodes in the Bayesian network candidate structures.   

With the feature strength indicator defined, our next step is to introduce an approach to building a candidate Bayesian network structure from data.  

\section{Method}

Formally, a Bayesian network is a pair $B = (G, P)$, where $G$ is a DAG that encodes a
joint probability distribution $P$ over a vector of random variables $\mathbf{F} = (f_1,\ldots, f_M)$ with
each node $n_i$ of the graph representing a variable in $\mathbf{F}$. The DAG can be
represented as a vector $G = (P_{n_1}, \ldots , P_{n_M})$ where each $P_{n_i}$
is a subset of the index set $V = \{ 1, \ldots, M \}$ and specifies the parents of node $n_i$ in the graph.

There are typically three types of nodes (features) $n_i$ in a Bayesian network which we will need to identify \textit{a priori} in our network generating algorithm
\begin{itemize}
    \item Parameters: These are the predictors for which we'd like to build the model. They have no dependencies, thus no children, and are often parameters such as age, gender/sex, Location/Geography, time/day, etc.
    \item Observed features: These are features which are observed and could have conditional dependencies, or be dependencies for other nodes.
    \item Unobserved features: These are features which are not directly observed and are being inferred from evidence or predicted. They could have conditional dependencies, or be dependencies for other nodes. These are typically labels or also could be latent variables. 
\end{itemize}

We now turn to the application of constructing Bayesian Network representations of data from the global features derived from weighted TMs. The approach is essentially a search problem that consists of two parts: scoring the top features for each targeted feature in the data using TM learning, and then a search algorithm that creates parent to child relationships based on most salient features found from the TMs. 

We assume a data set $\mathbf{X}$
comprised of a set of $M$ features $\mathbf{F} = {f_1, \ldots, f_M}$. Each feature $f_i$ we also assume can be represented by $d_{f_i}$ literals. For example, a feature with values low, medium, high would be represented by three literals $(l_1, l_2, l_3)$ with each $l_k \in (0,1)$. Low would be $(1,0,0)$, medium $(1,1,0)$, and high $(1,1,1)$.    
Starting with $f_1$, the goal is to traverse all variables of the data, where each variable is independently treated as an output variable of an independent TM model. For each variable, we learn which top $K$ features $f_{j}, f_{j+1}, \ldots, f_{j+K} \in \mathbf{F}$ have the most impact in predicting the variable as an output for a model. These top $K$ features are then proposed as candidate parent or child nodes of the targeted output variable.   

To do this we apply $R$ rounds of learning for each variable $f_i$ as an output target, and apply the global feature strength indicator \ref{global1} to all other features of the data set $X$. After each round, we aggregate all the feature strengths together and propose the top $K$ features as parent nodes of $f_i$.  

It is important to note that in learning a variable $f_i$, a balanced set of training data $X_I \subset X$ with regard to the output values of $f_i$ is helpful to ensure a fair assessment and reinforcement of literals for each feature. 

The construction of the Bayesian network candidate is done in three simple steps. The first step is to compute the most relevant features for all the the parameters, observed, and unobserved features in the the data.  Algorithm \ref{step1algo} demonstrates the first step in achieving a straightforward approach to compute a list of most relevant features for each type of variable $f_i$. Algorithm \ref{step2algo} will then apply the second step of the approach to each feature and relevant feature mapping.  

\begin{algorithm}
\caption{Learning the most salient global features for each variable}\label{step1algo}
\begin{algorithmic}[1]
\State Input $X$, $N$, $\epsilon$
\State Initialize an empty map $S_i : f \rightarrow \mathbb{Z}^{+}$ for each $f_i \in \mathbf{F}$ 
\For{each variable $f_i \in \mathbf{F}$}
\State Sample training set w.r.t outputs values of $f_i$
\State Learn $N$ rounds of $\epsilon > 0$ epochs \label{stateLearn} 
\For{each round in $N$}
\State Extract the top $K$ features and add to the map $S_i$
\EndFor
\EndFor
\State \textbf{return} $S_i$ 
\end{algorithmic}
\end{algorithm}

\begin{figure}[!ht]
\centering
\includegraphics[width=.45\textwidth]{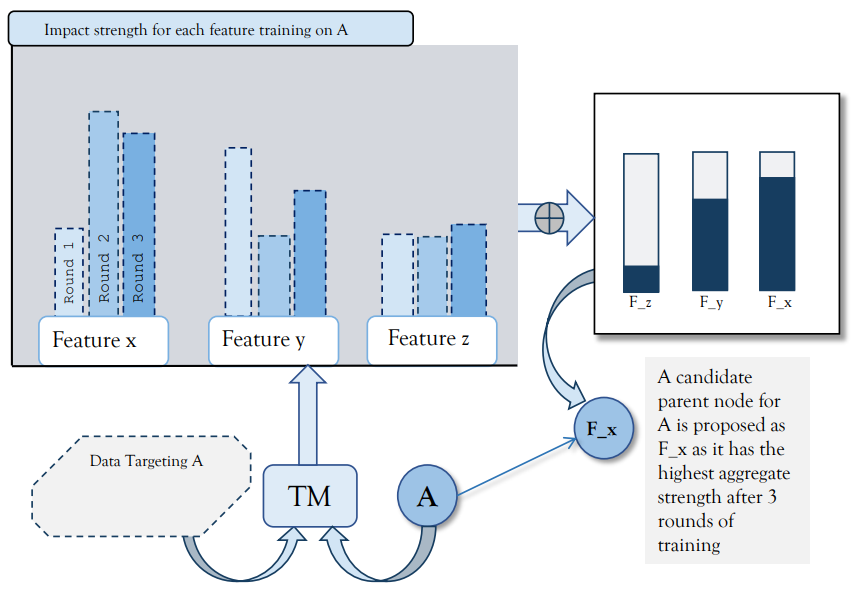}
\caption{Assembling the feature strengths for node (output) $A$. Most relevant feature $f_x$ from the strength map then proposed as parent node for $A$}
\label{ref:strength_to_node}
\end{figure}

Algorithm \ref{step1algo} begins by initializing a map for each variable $f_i$ that will be used to store the aggregation of the feature strengths $g[f_{\cdot}]$
when applying a learning round of a TM model.  Next, we traverse all the variables of the data set $X$, in no particular order, whereby fixing each variable $f_i$ as an 
output variable when applying the TM learning rounds. Each possible value of the the variable $f_i$ is thus treated as an output class $y = (y_1, y_2, \ldots, y_{d_f})$
where $d_f$ we recall is the dimension of the literal set $S_{f_i}$. In learning the output class given an in-sample training set of $X$, we essentially learn which features were
most salient when making a prediction for all output classes combined. The top $K$ features strengths $g[f_i]$ are added to the map as $S_i := S_i + g[f_i]$. After $M$ rounds 
of learning, we sort the aggregate feature strengths by rank of highest to lowest. These ranked features are candidates for parent nodes of $f_j$. To this end, we add the top $P$ features to a node list for feature $f_j$. This is represented in Figure \ref{ref:strength_to_node}, where we see that for a feature node (output) $A$, there are three features that provide high relevance $f_x$, $f_y$, $f_z$, and after three independent rounds of training, $f_x$ has the largest aggregate strength. We the propose the feature $f_x$ as a parent node for $A$. 

The output of Algorithm \ref{step1algo} is a map from each features $f_i$ to a list of the top $K$ features. For each feature, sorted by strength, the map will look like the table in \ref{topfeatureMap}, with $K=3$. 

\begin{table}
\centering
\caption{Top three features derived from each node}\label{topfeatureMap}
\begin{tabular}{|| l | c | c | c ||}
\hline
\textbf{Node} & TopFeature 1 & TopFeature 2 & TopFeature 3 \\
\hline 
$f_1$ & $f_{x}$ & $f_{y}$ & $f_{z}$  \\ 
$f_2$ & $f_{x_2}$ & $f_{y_2}$ & $f_{z_2}$  \\ 
$\dotsc$ & $\dotsc$ & $\dotsc$ & $\dotsc$  \\ 
$f_M$ & $f_{x_M}$ & $f_{y_M}$ & $f_{z_M}$  \\ 
\hline 
\end{tabular}
\end{table}

Before we introduce Algorithm \ref{step2algo}, we define the notion of \textit{lower strength direction}.
For an edge between two nodes $n_i$ and $n_j$ respective of features $f_i$ and $f_j$, the \textit{lower strength direction}
is the direction from the lower total feature strength to the higher. Namely, we sum up the feature strengths computed for each feature in regards to all the features, and the one with the lowest strength pointing in direction to the highest strength is the lowest strength direction. We will utilize this to remove competing edges between two nodes. 

\begin{algorithm}
\caption{Search Algorithm for Network Structure}\label{step2algo}
\begin{algorithmic}[1]
\Procedure{NetworkStructure}{ $S_i : f \rightarrow \mathbb{Z}^{+}$}
\State Empty set of nodes $\Phi_{X} = \{ \}$ 
\For{each "Parameter" node $n_i$}
\For{each TopFeature in list $S_i$}
\State Compute most frequent feature $f_{i,k}$ 
\State Define $f_{i,k}$ as parent node $n_j \leftarrow P_{n_i}$ 
\State Add $n_i$ and $n_j$ to $\Phi_{X}$  
\If{edge $n_i \rightarrow n_j$ exists}
\State remove edge
\EndIf
\EndFor
\EndFor
\For{each "Observed/Unobserved" node $n_i$}
\For{each TopFeature in list $S_i$}
\State Compute most frequent feature $f_{i,k}$ 
\State Define $f_{i,k}$ as child node $n_j \leftarrow P_{n_i}$
\State Add $n_i$ and $n_j$ to $\Phi_{X}$ 
\If{edge $n_j \rightarrow n_i$ exists} 
\State remove edge with lower strength direction
\EndIf
\EndFor
\EndFor
\EndProcedure
\end{algorithmic}
\end{algorithm}

Once the map table for each node is constructed, this algorithm is applied to build the first candidate network structure. It first passes over each parameter in the data $X$ independently of all the observed and any unobserved features. We find the features with the most impact in predicting that parameter and construct a parent node from the parameter to the global feature. After all the parameter features have been processed, we then proceed with finding any child nodes of the observable features. We then end with doing the same process with any unobserved features or labels/latent variables. 

In choosing each parent (for parameter nodes) or child node, beginning with the top feature, the algorithm begins by proposing the most frequently assigned top feature as the first node in the network. This feature being the most explanatory power for much of the data and thus a node and an edge is drawn between each of these pairings between the top feature node.  We then choose the second most frequented feature in the map of top features and propose it as a parent/child node, introducing edges between it and the nodes proposing that feature as a parent node. Each time we present a new edge, we must first verify that no cycle is introduced and that edge doesn't already exist. To ensure this, we measure the strengths of each feature relative to the node child node, and choose the edge which has a higher feature strength.

Continuing this until all top feature nodes are exhausted reveals the first candidate for the network. 
The final step in generating the network is to validate the network in that no cycles have been introduced. We again apply the lowest strength direction to ensure that maximal strengths on all edges remain. After Algorithm \ref{step2algo} has been applied, we apply a depth-first search algorithm to locate any cycles between 3 or more nodes. If a cycle is detected, we loop through all the edges in the cycle and remove the edge that has the lowest strength direction in the cycle that was introduced. Formally, this is given in Algorithm \ref{step3algo}

\begin{algorithm}
\caption{Cycle detection}\label{step3algo}
\begin{algorithmic}[1]
\State { $\Phi_{X}$, $e \in E$}
\For{each node $n_i \in \Phi$}
    \If{cycle of length $K$ exists}
        \State{$\phi \rightarrow \{n_i, n_{i,1}, \ldots, n_{i,K} \}$}
        \For{each possible edge in cycle $e \in \phi$}
            \If{if $e$ has minimum strength}
            \State{remove $e$ from $E$}
            \EndIf
        \EndFor
    \EndIf
\EndFor
\end{algorithmic}
\end{algorithm}

After removing any cycles from the network, a first generated DAG of the data has been produced. We can deduce from the three algorithmic steps that the following properties are ensured.
\begin{itemize}
    \item All designated parameters of the model have no dependencies (i.e. no child nodes) introduced. This is due to two properties designed in Algorithm \ref{step2algo}, namely 1) that parameter nodes only propose parent nodes from their feature strength list, and 2) that all other nodes will only propose all their feature nodes as child nodes (dependencies)
    \item All edges in the network are designated to find and connect conditional dependencies. This is due to the global interpretability property of TMs, that when learning on any labeled feature, the clauses will assemble literals from features that design patterns which are beneficial towards predicting that particular label value. 
    \item No cycles are introduced in the network. This is ensured by the last step \ref{step3algo} 
\end{itemize}

Before we begin our numerical examples and validation of the network generation, we first comment on some practical tips when applying the above algorithms.  
In practice, step \ref{stateLearn} in Algorithm \ref{step1algo} is quite crucial as it allows as to control the "information" that is distributed throughout the network. We will give strategies on this step in the following sections, but overall, we have found the best results when $N$, the number of iterations of the TM learning procedure, is quite large (say $N > 20$). Secondly, we also recommend using multiple TM models to boost robustness when building the map of important features. This will help ensure that the most crucial global features for each node are chosen. 

Now we move to the empirical section where we first validate our approach, and then offer two numerical examples on data where no known networks have been built before and the data comes from real-world industry scenarios.

\section{Car Insurance}\label{Insurance}

\begin{figure}[!ht]
\centering
\includegraphics[width=.4\textwidth]{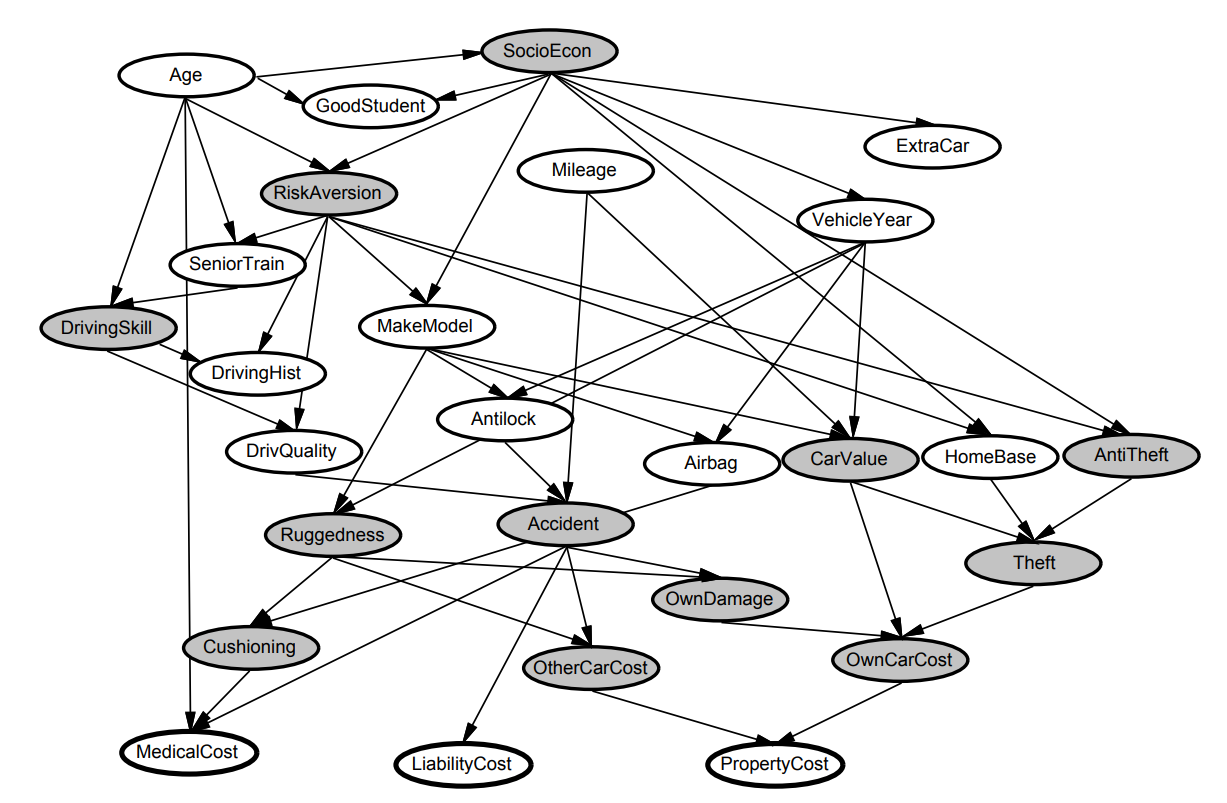}
\caption{Car insurance network to model expected claims costs}
\label{ref:insurance}
\end{figure}

To clearly demonstrate how the Algorithm \ref{step1algo} and Algorithm \ref{step2algo} builds a candidate Bayesian network model, in this section we will go over a full example where we have sampled data
directly from a well-known Bayesian network. The network is used for car insurance risk
estimation of the expected claim costs for a car insurance policy holder (see \cite{scutari2010learning} for more info on the model). Figure \ref{ref:insurance} shows the network which contains 27 nodes, of which 15 are observable, and over 1400 parameters. The hidden nodes (unobserved) are shaded and output nodes are shown with bold ovals. 

\begin{figure}[!ht]
\centering
\includegraphics[width=.4\textwidth]{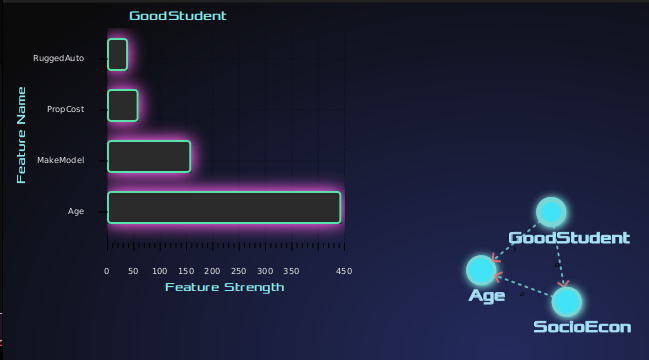}
\caption{Top features collected for predicting GoodStudent variable.}
\label{ref:goodStudent}
\end{figure}

We begin by simulating 5000 independent samples from the full joint probability definition of the network. For each variable,  we select a balanced training set with nearly identical samples of all outputs of that variable to ensure the most robust literal pattern design in the clauses as possible. In our first variable, GoodStudent, a two-level factor with levels False and True, following Algorithm \ref{step1algo} with 20 rounds of training, we've collected the top features and them in a list of candidate. Figure \ref{ref:goodStudent} showing the top features when predicting GoodStudent variable using the balance training set. Clearly, Age is shown to be the most prominent in this sample and is proposed as a parent node to GoodStudent. Next, we choose Age as the next node to model. After applying 5 rounds of learning, we find that RiskAversion is the feature with the most impact, 

\begin{figure}[!ht]
\centering
\includegraphics[width=.4\textwidth]{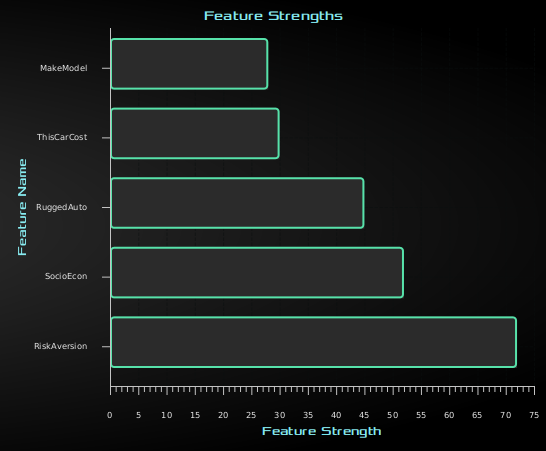}
\caption{Top features collected for predicting Age variable.}
\label{ref:ageEstimation}
\end{figure}

Continuing along, very every node, the top 5 features are computed using 20 rounds of learning for the next several nodes. RiskAversion having PropCost, Age, and CarValue as top parent candidates. The table in \ref{topfeatures} shows in order of computation, the top 3 features for every node which completes the first stage in building a candidate Bayesian network structure. 

\begin{figure}[!ht]
\centering
\includegraphics[width=.4\textwidth]{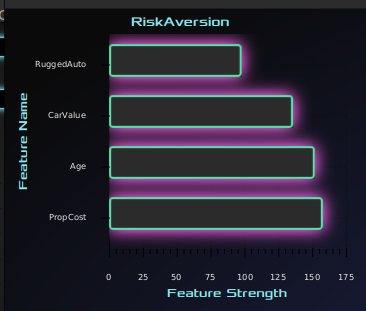}
\caption{Showing the global feature strengths for RiskAversion}
\label{fig:example2}
\end{figure}

In the Top Feature columns, we use bold font to highlight that the top feature node is a child of the targeted node, namely that the targeted node has dependencies in the TopFeatures and italic font to denote that they are an ancestor. 

Comparing the table of the top features chosen for each node with the original Bayesian network, we see further details in the accuracy of choosing potential parent node candidate. Especially for nodes with two or more parent nodes, we notice high accuracy in choosing those nodes as top features after several rounds of TM learning. For example, SeniorTrain has RiskAversion and Age as two parent nodes, and were computed as the top two features. OtherCar has SocioEcon as a parent node and is also a top feature.  In the case a node has no parents, namely, it's a prior probability that is used for connected nodes, the feature strength selection process yields similar results, but not as robust as with nodes predicting their parent nodes.

For example, Age is the parent node providing observable prior information for two of its top features. ThisCarCost is the only node which had top features of which none were either true parent nodes of grandparent nodes in the underlying network, nor were child nodes for any feature. SocioEcon, with MakeModel and RiskAversion as its top two features, were not parent nodes in the underlying network, but were child nodes, thus having immediate dependencies for those nodes. 

\begin{table}
\centering
\caption{Top three features derived from each node}\label{topfeatures}
\begin{tabular}{|| l | c | c | c | c ||}
\hline
\textbf{Node} & TopFeature 1 & TopFeature 2 & TopFeature 3 \\
\hline
SeniorTrain & \textbf{RiskAversion} & \textbf{Age} & RuggedAuto  \\ 
OtherCar & \textbf{SocioEcon} & RiskAversion & Mileage  \\ 
Antilock & CarValue & \textbf{MakeModel} & SocioEcon  \\ 
ILiCost & \textbf{Accident} & MakeModel & \textit{DrivQuality}  \\ 
Cushioning& \textbf{RuggedAuto} & RiskAversion & SocioEcon  \\ 
CarValue& \textbf{MakeModel} & \textbf{Mileage} & RiskAversion    \\ 
PropCost& RiskAversion & Age & DrivQuality  \\ 
Airbag & \textbf{MakeModel} & RuggedAuto & Accident   \\ 
Theft & \textit{RiskAversion} & \textit{SocioEcon} & \textit{Mileage}  \\ 
ThisCarDam & \textbf{Accident} & DrivQuality & RiskAversion    \\ 
DrivHist & \textbf{RiskAversion}  & DrivQuality & \textbf{DrivingSkill}  \\ 
ThisCarCost & DrivQuality & Age & RiskAversion  \\ 
RuggedAuto& \textbf{MakeModel} & \textit{SocioEcon} & \textit{RiskAversion}  \\ 
VehicleYear & CarValue & ThisCarCost & \textbf{SocioEcon}  \\ 
DrivingSkill& DrivQuality & MakeModel & \textit{SocioEcon}  \\ 
GoodStudent& \textbf{Age} & MakeModel & RiskAversion  \\ 
DrivQuality& \textbf{RiskAversion} & MakeModel & Mileage  \\ 
\texttt{Age} & RiskAversion & Theft & Mileage  \\ 
HomeBase& \textbf{SocioEcon} & \textbf{RiskAversion} & \textit{Age} \\ 
Accident & \textbf{DrivQuality} & \textbf{Mileage} & \textit{RiskAversion}  \\ 
SocioEcon & MakeModel & RiskAversion & Theft  \\ 
MedCost& \textit{RuggedAuto} & RiskAversion & SocioEcon   \\ 
\texttt{Mileage} & Age & MakeModel & RiskAversion  \\ 
MakeModel& \textbf{RiskAversion} & Mileage & RuggedAuto  \\ 
AntiTheft& \textbf{RiskAversion} & \textbf{SocioEcon} & Mileage  \\ 
RiskAversion& \textbf{Age} & MakeModel & Mileage  \\ 
OtherCarCost & \textbf{RuggedAuto} & MakeModel & RiskAversion  \\ 
\hline
\end{tabular}
\end{table}

With the table of nodes their top features estimated, we can now apply algorithm \ref{step2algo} that essentially proposes the child or parent node candidates and then prunes the network. The second pass of the algorithm builds the structure while also respecting both parameter and observed node types. We begin by choosing the two parameters Age, and Mileage and find the parent nodes by taking the TopFeature, in this case RiskAversion for Age which suggests that RiskAversion was typically the most prominent feature in predicting the age of the driver given all other 26 variables.  

\begin{figure}[!ht]
\centering
\includegraphics[width=.4\textwidth]{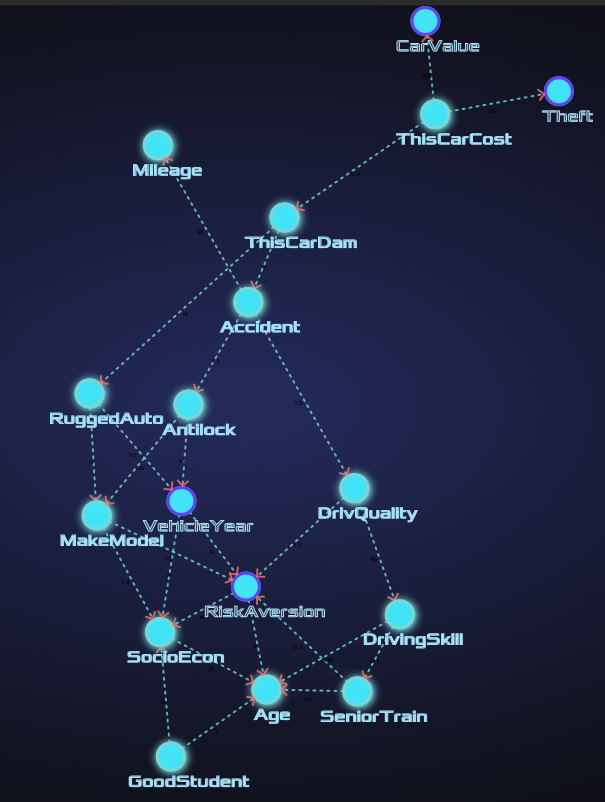}
\caption{First round of the algorithm yields a network close to ground truth}
\label{ref:buildOne}
\end{figure}

Next we consider our observed variables, in this case GoodStudent, followed by choosing the first top feature which in this case is Age. With the next candidate node chosen as Age, the algorithm then selects the top feature extracted RiskAversion.  Now selecting RiskAversion as the next candidate node, we see that Age is selected as the top feature. which creates a scenario in which a root node is also a child node. In the case of an edge introduced in the opposite direction, the higher feature is chosen, and in this case RickAversion is directed towards Age as it has the higher feature direction. Thus Age becomes the parent node for RiskAversion. This is shown in Figure \ref{ref:buildOne}. With the first three nodes being introduced, GoodStudent, Age, and RiskAversion with Age determined as the parent for both GoodStudent and RiskAversion, the next step is to introduce a new node candidate randomly, SocioEcon. Computing the most prominent feature yields MakeModel with a second being RiskAversion. 

Figure \ref{ref:buildOne} depicts the candidate network after a first round of the \ref{step1algo}. Notice that Age is the dependency on many nodes at the bottom of the network structure. This corresponds to ground truth network, except for the missing of several parameter nodes which we introduce next. 

Completing the Algorithm \ref{step2algo} over all observed nodes, we arrive to a network close to ground truth, with the only additional edges on the graph being AntiLock being partially dependent on CarValue and Mileage being partially dependent on Age. Both of these edges could be argued to be included in the final network, but as we are wanting to see how close we are to the original network from which we simulated the data, we will need an approach to compare "distance" between graphs. The best method for this is in so called "graph kernels" which are functions that compute an inner product on graphs. 

\subsection{Comparison}

To compare the proximity to the ground truth network, we apply a kernel graph based approach to compare a TM generated network with the proposed network from which the underlying data was simulated. We would like to understand how similar two nodes are in the graph, comparing both outgoing edges and incoming edges, as well as how similar the node's neighbors are. 

Graph kernels are used widely in many disciplines and typically follow three steps, which we will use in our comparison study. For more on graph kernels, \cite{vishwanathan2008graph} contains an in-depth study with various applications.  

\begin{itemize}
\item  \textbf{Compute Node Similarity}: This step computes the pairwise similarity between nodes in two graphs, iterating over all nodes in both graphs and calculating the similarity between corresponding nodes.
\item  \textbf{Compute Edge Similarity}: Computes the pairwise similarity between edges in the two graphs similar to node similarity, it involves iterating over all edges in both graphs and calculating the similarity between corresponding edges and their direction.
\item  \textbf{Combine Similarity}: Lastly, we combine the node similarity matrix and edge similarity matrix into an overall graph similarity score by averaging the similarities.
\end{itemize}

Since the graph kernel computation depends on the number of nodes and edges of a graph, we normalize the values of the averaging of similarities to get a value between 0 and 1, with 0 being the value comparing a null graph with any non-empty graph, and 1 being all the nodes and edges with their directions are identical.  

We apply our network structure generating algorithm to four different popular networks. In order to create a robust framework for the comparison, we run our algorithm using 5 different randomly chosen Tsetlin Machine parameterizations. We do this to remove any chance that we could be "overfitting" our models for generating the networks. We use a simple strategy of randomly choosing the number of clauses to be a random integer between 1 and 3 times the total number of literals for the data simulated by the network. The threshold parameter is then chosen as a random number between 10 and the number of clauses, coupled with the specificity between 5 and 15. Finally, the maximum number of literals per clause is chosen to be a random value between 3 and the total number literals for the data. Table \ref{Collections} shows the randomly chosen parameterizations, with $L$ being the total number of literals for the data.  

\begin{table}
\centering
\caption{Five models used to combine into ensemble model}\label{Collections}
\begin{tabular}{|| l | c | c | c | c ||}
\hline
  & Clauses & Threshold & Specificity & Max Number Literals  \\
\hline
Model I & L + 12 & 25 & 9.4 & 25  \\
\hline
Model II & L + 34 & 21 & 11.2 & 33 \\
\hline
Model III & L + 23 & 30 & 6.7 & 7  \\
\hline
Model IV & L + 58 & 10 & 9.8 & 55  \\
\hline
Model V & L + 92 & 42 & 12.6 & 5  \\
\hline
\end{tabular}
\end{table}

\begin{figure}[!ht]
\centering
\includegraphics[width=.5\textwidth]{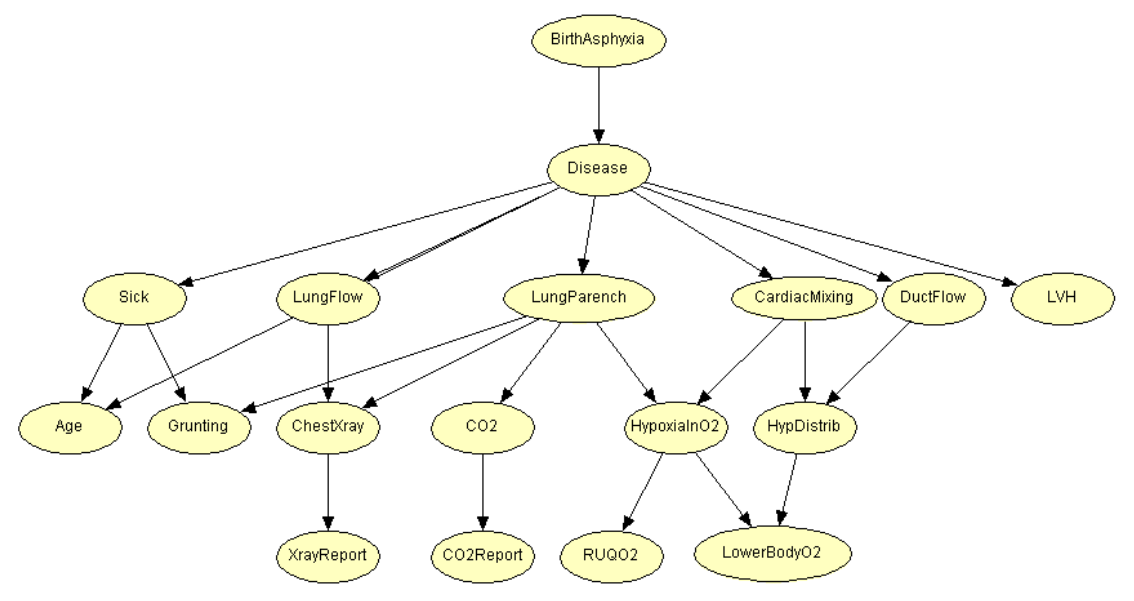}
\caption{Child network, adapted from \cite{scutari2010learning}}
\label{ref:child}
\end{figure}

The four networks we simulate data from are as follows:
\begin{itemize}
    \item Asia: A small simplified version of a network that may be used to diagnose a doctor's new patients. Each node represents a facet relating to a patient's condition, and each directional edge roughly corresponds to causality.
    \item Child: A larger network modeling uncertainty of birth asphyxia given lung metrics, CO2, x-rays, and other variables (see \ref{ref:child})
    \item Insurance: The network from section \ref{Insurance}
    \item Diabetes: A large network used for the detection of early signs and symptoms of diabetes. Such models are important to control the side effects of diabetes and Bayesian network models have been used extensively in the literature.     
\end{itemize}

The strategy for our study is to generate 10 networks from each model, and compute the similarity score using the kernel graph similarity computation. The average of these scores is then recorded for each model. Table \ref{scoresAll} shows the results for the four different networks. 

\begin{table}
\centering
\caption{Four networks and average similarity score across fives models}\label{scoresAll}
\begin{tabular}{|| l | c | c | c | c | c |}
\hline
Network & Model I & Model II  & Model III & Model IV & Model V \\
\hline
Asia & 0.834 & 0.855 & 0.791 & 0.846 & 0.821 \\
\hline
Child & 0.875 & 0.891 & 0.908 & 0.940 & 0.805 \\
\hline
Insurance & 0.831 & 0.874 & 0.924 & 0.981 & 0.819 \\
\hline
Diabetes & 0.754 & 0.790 & 0.781 & 0.877 & 0.825 \\
\hline
\end{tabular}
\end{table}

We see that typically, the medium (20 - 50) node networks tend to do best in performance on average. The smallest network had the worse performers in terms of models, with the Child and the Insurance networks the best performers across two of the parameterizations. The very large network (> 100 nodes) performs well on high specificity with high amount of literals allowed per clause. It seems that on average, high specificity parameter coupled with allowing more literals per clause seems to work better than being more restrictive on the number of literals per clause (models III and V). This is mostly likely because for very challenging target nodes, allowing more information via literals in the clauses allows for more complex patterns. 

While generating networks based on a ground-truth model plays an important role in this study, we would also like to see the performance of the network generating algorithm using data that is unlabeled and has no ground-truth for reference. Here we rely on intuition and logic along with ensuring the networks generated satisfy the properties of a DAG to derive how well the algorithm does. 

\section{Supply Chain Analysis}

In our final example, we consider a supply chain data set collected from a Fashion and Beauty startup that is based on the supply chain of makeup products \cite{supplychain}. Supply chain analytics is the process of collecting, modeling, and interpreting data related to the movement of products and services from suppliers to customers.  

The challenge with this particular data set is that is it quite small: only 100 observations from the supply chain exist, namely one data point for each product. With 23 different variables accounting for the supply chain for each of the products, creating a predictive model for effectively understanding the sensitivities of revenue generated, the impact of how stock and MC attribute to profits, and many of the other variables could be challenging for traditional models such as gradient boost, neural network, or random forests where larger data sets are typically required. Using our TM formulation for building network structures of the data, we will see that even though we only have one observation for each product, achieving a coherent network structure is still quite readily obtainable.

\begin{figure}[!ht]
\centering
\includegraphics[width=.5\textwidth]{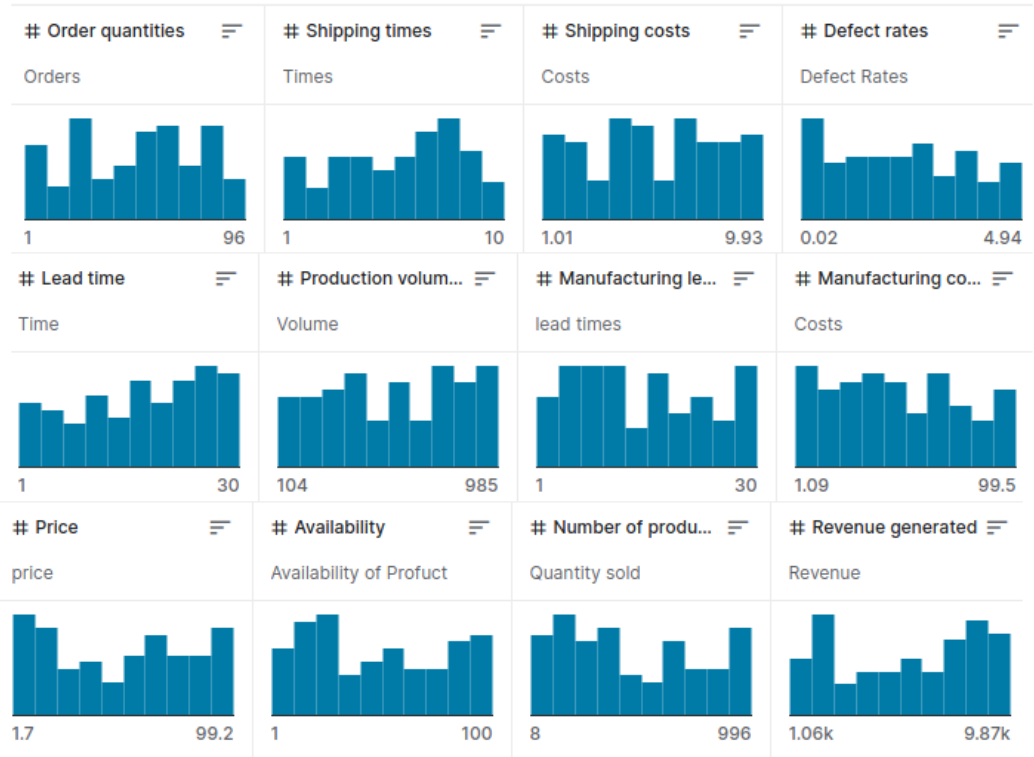}
\caption{Continuous features of the supply chain data set}
\label{ref:allVar}
\end{figure}

\begin{figure}[!ht]
\centering
\includegraphics[width=.5\textwidth]{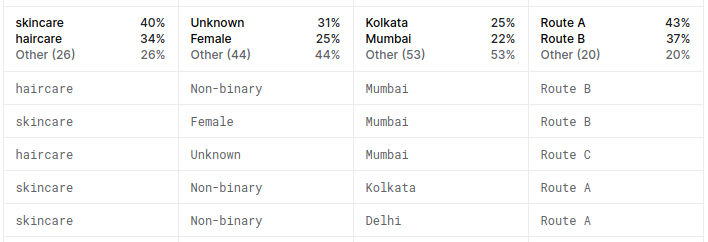}
\caption{Some categorical features of the supply chain data set}
\label{ref:categorical}
\end{figure}

Figure \ref{ref:allVar} shows some of the major continuous variables featured in the supply chain data set along with their value frequency. The easiest way to transform the continuous variables into observations easily handled by both TMs and Bayesian networks is to simply attribute the values to \verb|low, medium, high, very high|. For the categorical variables, such as the ones shown in figure \ref{ref:categorical}, we simply find the frequency each variable and rank them from least frequency to most frequent. The least frequent value will be a 1, second most frequent a 2, and so on.  

With so few observations, obtaining near optimal hyper parameters for our TM model will be challenging. We propose a simple way around this by building an ensemble model, which is a collection of smaller models each with different number of clauses, thresholds, and specificity values. Our global feature impact model will then be an aggregation of all the literals in each clause in each model. 

We begin by applying our ensemble learning approach on each of the discretized continuous variables. For this, we set the variable as the target label, thus giving four different values to learn on \verb|low, medium, high, very high|.

\begin{table}
\centering
\caption{Five models used to combine into ensemble model}\label{EnsembleModel}
\begin{tabular}{|| l | c | c | c | c ||}
\hline
  & Clauses & Threshold & Specificity & Max Number Literals  \\
\hline
Model I & 42 & 14 & 7.2 & 18  \\
\hline
Model II & 80 & 37 & 7.4 & 32  \\
\hline
Model III & 66 & 35 & 8.7 & 27  \\
\hline
Model IV & 85 & 10 & 9.4 & 85  \\
\hline
Model V & 42 & 35 & 11.3 & 42  \\
\hline
\end{tabular}
\end{table}

Being the first variable we target, is a variable that is most likely determined at the origin of the supply chain. Applying our 5 independent TM models to learn the high impact features, we discover that \verb|Availability|, and \verb|Production volumes| are the two features which score very high in prediction impact. Namely, to most accurately predict price, these two variables have the most impact. A third is \verb|Stock levels| which we will note in our node to feature table \ref{topfeaturesSupply}. 

With the highest impact being \verb|Availability|, we now target this variable to learn its most salient features. After one round of training with each of the 5 models, we see that \verb|Order quantities| and \verb|Defect rates| play the largest roles, with \verb|Manufacturing costs| a close third. We then consider \verb|Order quantities|, and see that 
\verb|Stock levels| and \verb|Price| are the highest predictors with \verb|Revenue generated| very strong as well. 

Continuing through the rest of the variables, we finalize our node to feature table \ref{topfeaturesSupply} and can start building an initial directed network where each node is a variable from the supply chain.

\begin{table}
\centering
\caption{Top three features derived from each node}\label{topfeaturesSupply}
\begin{tabular}{|| l | l | l | l ||}
\hline
\textbf{Node} & TopFeature 1 & TopFeature 2 \\
\hline
Price & Availability & ProdVolumes \\ 
\hline
Availability & Order quantities & Defect rates  \\
\hline
Order quantities & Stock levels & Price  \\
\hline
Lead times & Lead time & Stock levels  \\
\hline
Shipping times & MLT & Lead time  \\
\hline
Stock levels & Lead times & Availability  \\
\hline
Shipping carriers & Order quantities & Shipping times \\
\hline
Customer demographics & RevGenerated & Location \\
\hline
RevGenerated & MC & Stock levels \\ 
\hline
NumProdSold & MC & Defect rates   \\
\hline
Product type & Lead temp & Availability  \\
\hline
Supplier name & Availability & Location  \\
\hline 
Shipping costs & Lead time & Order quantities  \\
\hline
Location & Price & ProdVolumes  \\
\hline
Lead time & Availability & ProdVolumes  \\
\hline
ProdVolumes & NumProdSold  & Stock levels   \\
\hline
MLT & NumProdSold & Lead time \\
\hline 
MC & Order quantities & MLT  \\
\hline
Inspection results & Defect rates & RevGenerated  \\
\hline
Defect rates & RevGenerated  & Order quantities    \\
\hline
Transportation modes & Shipping times & Stock levels  \\
\hline
Routes & Shipping times & Price  \\
\hline
Costs & Order quantities  & Defect rates   \\
\hline

\end{tabular}
\end{table}

\begin{figure}[!ht]
\centering
\includegraphics[width=.5\textwidth]{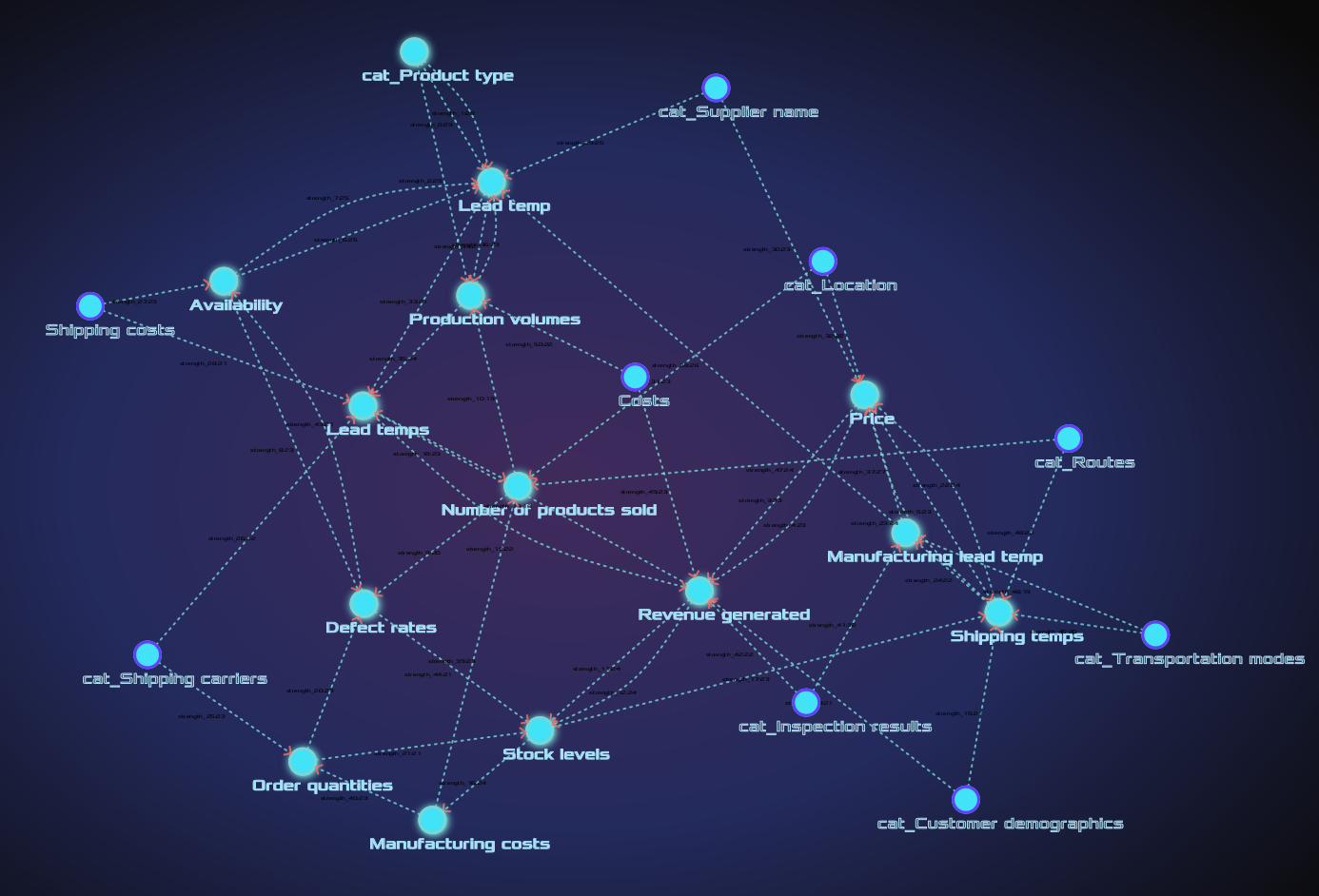}
\caption{First iteration of the network using top two features for each node}
\label{ref:supplynet}
\end{figure}

In our first iteration of the Algorithm \ref{step1algo}, a collection of vertices and edges are produced. The second iteration, Algorithm \ref{step2algo} runs over the network and looks for competing edges and eliminates any edges with a lower strength direction. This gives  the final network as shown in Figure \ref{ref:supplynet}. To ensure there are no cycles in the graph, we do a depth first search on all the nodes. In any case there is a cycle, we simply eliminate the edge with the lowest strength direction in the cycle.    

In analyzing the final network, we see that the TM to BN algorithm has discovered that Product type, Shipping costs, Shipping carriers, Routes, Customer Demographics, and Location are some of the main input parameters into the supply chain. This seems to be consistent with the logistics of supply chains and how they operate in the sense that they are degrees of freedom that can be used to optimize costs in the supply chain. Namely, they will have an impact on costs, maintenance, stock, and lead times. 

The direct conditionals on Revenue generated has been discovered to be \verb|Price|, \verb|Costs|, \verb|Number of Products sold|, \verb|Stock levels| and \verb|Inspection Results|,  which from a quick sanity check registers well with typical revenue models. A failed \verb|Inspection Results| could lead to higher costs, or manufacturing costs which would impact the revenue generated.  

Lastly we see that another important observable node in the supply chain \verb|Number of products sold| is dependent on \verb|location|, \verb|Production Volumes| and \verb|Lead times|. In return, it is computed that \verb|Number of products sold| influences the \verb|defect rate| as perhaps defect rates scale with the number of customers using the product.  

We have also shown two competing directions between the \verb|Availability| variable and \verb|defect rate|. Here, the feature strength in the direction from \verb|defect rate| to \verb|Availability| is significantly higher than the other direction, making the availability of the product conditionally dependent on the defect rate. 

On a final note, we mention that although we now seem to have a candidate for a network structure, a fully constructed Bayesian network still requires conditional probability tables across all the parent nodes and their children. If we do not have expert knowledge into these probability distributions, the easiest way is to estimate them empirically. With the network proposal built, generating conditional probability tables for a particular variable node $v$ can be done simply by computing the frequency of each child node's literals of $v$, and ensuring they are normalized across all combinations of literals for that child. 

\section{Bank Customer Churn}

In this last example we will build a candidate network for Bank Customer Churn data. The data in this example is the customer data of account holders from an Anonymous Multinational Bank and the aim of the data is to not only predict the customer churn, but understand when predicted, why does it occur. Every bank wants to hold there customers for sustaining their business for a long period of time. But as customers might leave their banking institution for a variety of reasons, generating a Bayesian network for gaining better insights into customer retention could be useful, e.g. when a customer has a high probability of leaving given their current conditions. Reasons why a banking institution might want to minimize customer turnover is because it is typically much more expensive to sign in a new client than keeping an existing one.
It is advantageous for banks to know what leads a client towards the decision to leave the company. Churn prevention allows companies to develop loyalty programs and retention campaigns to keep as many customers as possible.

The data contains 15 features, of which 4 are parameters, 10 observable, and 1 label (namely whether the customer has left bank). The features are as follows:

\begin{itemize}
    \item CreditScore: can have an effect on customer churn, since a customer with a higher credit score is less likely to leave the bank.
    \item Geography: a customer’s location can affect their decision to leave the bank.
    \item Gender: it’s interesting to explore whether gender plays a role in a customer leaving the bank.
    \item Age: this is certainly relevant, since older customers are less likely to leave their bank than younger ones.
    \item Tenure: refers to the number of years that the customer has been a client of the bank. Normally, older clients are more loyal and less likely to leave a bank.
    \item Balance: also a very good indicator of customer churn, as people with a higher balance in their accounts are less likely to leave the bank compared to those with lower balances.
    \item NumOfProducts: refers to the number of products that a customer has purchased through the bank.
    \item HasCrCard: denotes whether or not a customer has a credit card. This column is also relevant, since people with a credit card are less likely to leave the bank.
    \item IsActiveMember: active customers are less likely to leave the bank.
    \item EstimatedSalary: as with balance, people with lower salaries are more likely to leave the bank compared to those with higher salaries.
    \item Exited: whether or not the customer left the bank.
    \item Complain: customer has complaint or not.
    \item Satisfaction Score: Score provided by the customer for their complaint resolution.
    \item Card Type: type of card hold by the customer.
    \item Points Earned: the points earned by the customer for using credit card.
\end{itemize}

We can identify already the three parameters (predictors) being Geography, Gender, and Age as they have no dependencies. All other features clearly have some kind of partial dependencies, or might also supply conditions for other features.  
For the observed variables, we look first at CreditScore, EstimatedSalary, and SatisfactionScore which have a few very strong partial dependencies in predicting churn as we'll see. Table \ref{topfeaturesBank} shows the top features of each observed feature.  

\begin{table}
\centering
\caption{Top three features derived from each node}\label{topfeaturesBank}
\begin{tabular}{|| l | l | l ||}
\hline
\textbf{CreditScore} &  \textbf{EstimatedSalary} & \textbf{SatisfactionScore} \\
\hline
Age  & Geography & NumOfProducts \\
Tenure & Gender & PointEarned \\
EstimatedSalary & Tenure & EstimatedSalary \\
\hline
\end{tabular}
\end{table}

We see that CreditScore has dependencies on Age and Tenure, which is typically in line whith how credit models work for many retail banking institutions. EstimatedSalary is geographically dependent as some countries have higher economic power. For the Satisfaction Score, it seems that it is most dependent on NumOfProducts a customer has purchased through the bank, and how often they use their credit card.

\begin{figure}[!ht]
\centering
\includegraphics[width=.4\textwidth]{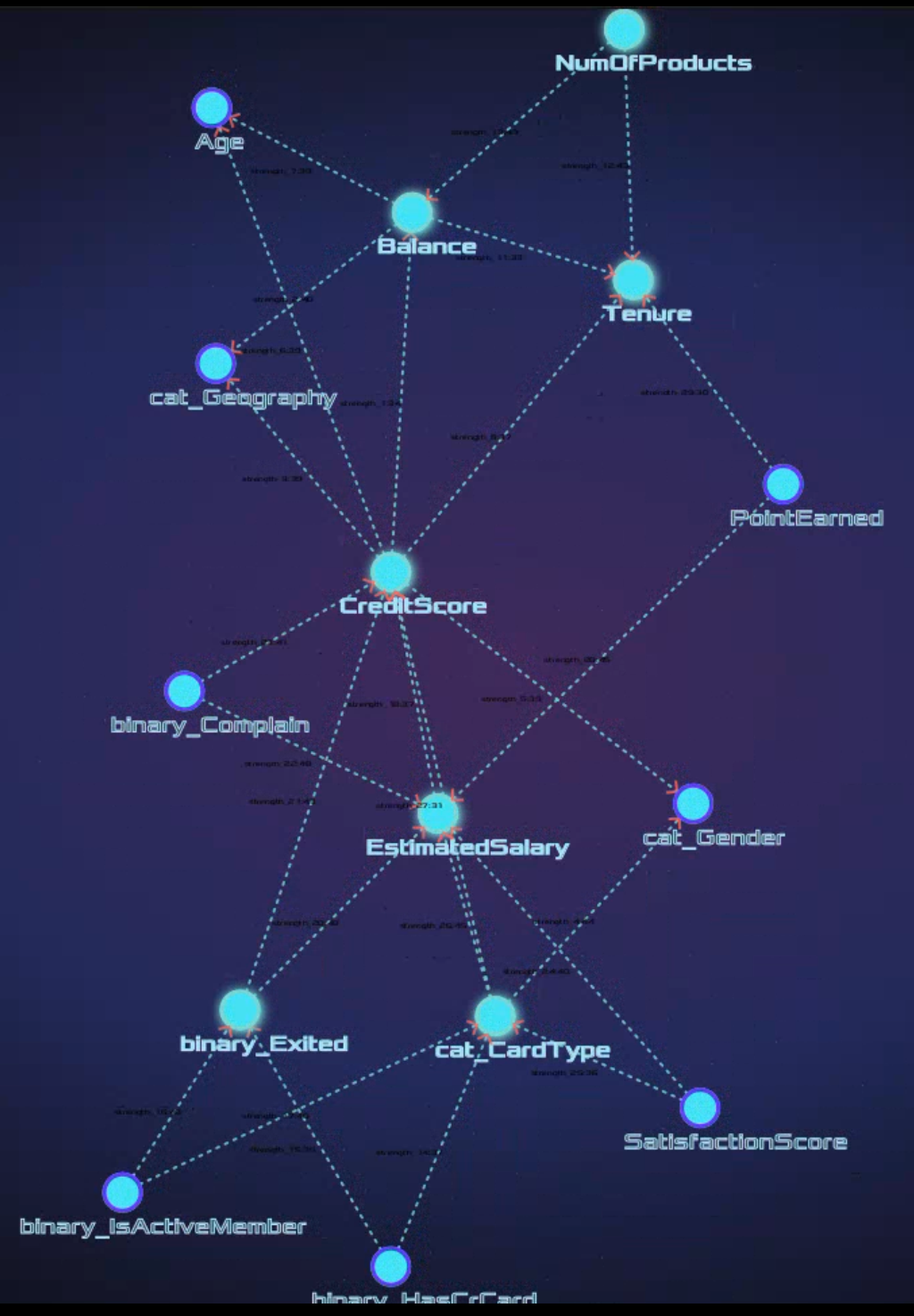}
\caption{First candidate network for Bank Customer Churn data}
\label{ref:bankChurn}
\end{figure}

Our first candidate network for the Bank customer churn data is shown in Figure \ref{ref:bankChurn}. We see that the label (unobserved) node Exited, namely whether the customer will leave the banking institution is mostly dependent on their CreditScore and EstimatedSalary. In turn, these features are highly dependent on Balance in the account, Age, and Tenure at the institution. This makes sense in most models for customer churn as normally, older clients are more loyal and less likely to leave a bank.

\section{Conclusion}

We proposed an algorithm for constructing a Bayesian network using feature strength indicators derived from the clause weights and literals of a Tsetlin machine. We showed through both empirical evidence in comparing with ground-truth networks as well as through examples where dependencies on certain features in the data are widely know that TMs generating initial networks through data is a very promising area of research.  The strengths are numerous, which one being that the method is entirely interpretable which how parent and child nodes are generated, and also in the shear speed of the approach - candidate networks can be generated in milliseconds to seconds, even for large data sets with many features (~100). 

While this first proposal is not without drawbacks or challenges, it has shown to be an attractive basis for further research and methodical development. One area that seems like a reasonable direction would be in a hybrid-based structure learning. Consider for example a score based approach such as in the Chow-Liu algorithm \cite{scutari2010learning} where instead of mutual information is used to construct a spanning tree, feature strengths derived from TMs are used instead to construct the first sequence of trees. 

Another drawback is the fact that a good foundation for generating a network should start with knowing the parameters of the model being generated, namely which features of the data do not have any dependencies. This can be challenging for many large data sets without expert witnesses. Furthermore, this approach cannot introduce non-labeled latent variables as all the features that are candidate nodes need to be present in the data.

We demonstrated that building a Bayesian network with TMs can also reveal deeper interpretable insights into the data. In our Supply Chain example, even with only 100 data observations from a supply chain, we were able to build a network with a good understanding of the inter dependencies in the network. For the Bank Customer Churn, our network generating algorithms were able to create a network that fit quite well the narrative to how customer exit models behave in general. 

Our future work will be in several areas in improving the approach. First, we would like to improve on algorithms \ref{step1algo} and \ref{step2algo} in both filtering for Top Features in a more constructive way by perhaps working with a newly developed Type-III feedback system (paper to appear). This method seeks to remove in a more direct way literals from features which do not add any value to clause feedback and thus predictions overall. The idea here would be that more literals would be penalized and thus the final Top Features would have truly more impact. The second area is to better design how competing directions in the network are resolved. Here we simply take the higher feature strength between two competing connections. 

Lastly, our next applications of interest will be in designing temporal/dynamic networks on time-dependent data, where nodes can be connected via lags in the data. 

\bibliographystyle{named}
\bibliography{bn23}

\appendix \label{appendix:weighted}

In this subsection, we give some more details of the TM with weights model. The learning of weights is based on increasing the weight of clauses that receive true positive output (Type Ia feedback) and decreasing the weight of a clause that receive false positive outputs (penalizing Type II feedback). The motivation is to determine which clauses are inaccurate and thus must team up to obtain high accuracy as a team (low weight clauses), and which clauses are sufficiently accurate to operate more independently (high weight clauses). The weight updating procedure is summarized in Algorithm \ref{algo:tm}. Here, $w_i$ is the weight of clause $C_i$ at the $n^{th}$ training round (ignoring polarity to simplify notation). The first step of a training round is to calculate the clause output. The weight of a clause is only updated if the clause output $C_i$ is 1 and the clause has been selected for feedback ($P_i$ = 1). Then the polarity of the clause and the class label $y$ decide the type of feedback given. That is, like a regular TM, positive polarity clauses receive Type Ia feedback if the clause output is a true positive, and similarly, they receive Type II feedback if the clause output is a false positive. For clauses with negative polarity, the feedback types switch roles. When clauses receive Type Ia or Type II feedback, their weights are updated accordingly. We use the stochastic searching on the line (SSL) automaton to learn appropriate weights. SSL is an optimization scheme for unknown stochastic environments~\cite{oommen}. The goal is to find an unknown location $\lambda^*$ within a search interval $[0,1]$. In order to find $\lambda^*$, the only available information for the Learning Mechanism (LM) is the possibly faulty feedback from its attached environment $E$.\\
In SSL, the search space $\lambda$ is discretized into $N$ points, $\{0,1/N,2/N,...,(N-1)/N,1\}$ with N being the discretization resolution. During the search, the LM has a location $\lambda \in \{0,1/N,2/N,...,(N-1)/N,1\}$, and can freely move to the left or to the right from its current location. The environment $E$ provides two types of feedback: $E = 1$ is the environment suggestion to increase the value of $\lambda$ by one step, and $E = 0$ is the environment suggestion to decrease the value of $\lambda$ by one step. The next location of $\lambda$, i.e. $\lambda_{n + 1}$, can thus be expressed as follows:
\begin{equation}
 \lambda_{n + 1} =
    \begin{cases}
      \lambda_n+1/N, & \text{if $E_n=1$,}\\
      \lambda_n-1/N, & \text{if $E_n=0$.}\\
    \end{cases}       
\end{equation}
\begin{equation}
 \lambda_{n + 1} =
    \begin{cases}
      \lambda_n, & \text{if $\lambda_n=1$ and $E_n=1$,}\\
      \lambda_n, & \text{if $\lambda_n=0$ and $E_n=0$.}\\
    \end{cases}       
\end{equation}
Asymptotically, the learning mechanics is able to find a value arbitrarily close to $\lambda^*$ when $N\rightarrow\infty$ and $n\rightarrow\infty$. In our case, the search space of clause weights is $[0, \infty]$, so we use resolution $N = 1$, with no upper bound for $\lambda$. Accordingly, we operate with integer weights. As in Algorithm~\ref{algo:tm}, if the clause output is a true positive, we simply increase the weight by $1$. Conversely, if the clause output is a false positive, we decrease the weight by $1$.\\
By following the above procedure, the goal is to make low precision clauses team up
by giving them low weights, so that they together can reach the summation target $T$. By teaming up, precision increases due to the resulting ensemble effect. Clauses with high precision, however, obtain a higher weight, allowing them to operate more independently.\\
The above weighting scheme has several advantages. First of all, increment and decrement operations on integers are computationally less costly than multiplication based updates of real-valued weights. Additionally, a clause with an integer weight can be seen as multiple copies of the same clause, making it more interpretable than real-valued weighting, as shown in the next section. Additionally, clauses can be turned completely off by setting their weights to $0$ if they do not contribute positively to the classification task. For a more detailed explanation of the weighted TM, please refer to ~\cite{abeyrathna2021integer}.

\begin{algorithm}[t]
\caption{Complete WTM learning process}\label{algo:tm}
\begin{algorithmic}[1]
\State \textbf{Input:} Training data batch $(B, x, y) \quad \rhd B \geq 1$
\State \textbf{Initialize:} Random initialization of TAs
\State \textbf{Begin:} $n^{th}$ training round
\For{$i = 1, ...,m$} \textbf{if} $P_i = 1$  
    \If{($y = 1$ \textbf{and} $i$ is odd) \textbf{or} ($y = 0$ \textbf{and} $i$ is even)}
        \If{$c_i = 1$} 
            \State $w_i \leftarrow w_i+1$ 
            \For{feature $k=1,...,2o$}
                \If{$l_k = 1$}
                    \State Type Ia Feedback
                \Else:
                    \State Type Ib Feedback
                \EndIf
            \EndFor
        \Else:
            \State $w_i \leftarrow w_i \quad \rhd$ \text{[No Change]}
            \State Type Ib Feedback
        \EndIf
    \Else: ($y = 1$ \textbf{and} $i$ is even) \textbf{or} ($y = 0$ \textbf{and} $i$ is odd)
        \If{$c_i = 1$} 
            \If{$w_i > 0$}
                \State $w_i \leftarrow w_i-1$ 
            \EndIf
            \For{feature $k=1,...,2o$}
                \If{$l_k = 0$}
                    \State Type II Feedback
                \Else:
                    \State Inaction
                \EndIf
            \EndFor
        \Else:
            \State $w_i \leftarrow w_i \quad \rhd$ \text{[No Change]}
            \State Inaction
        \EndIf
    \EndIf
\EndFor
\end{algorithmic}
\end{algorithm}

\end{document}